\documentclass[letterpaper, 10 pt, conference, final]{ieeeconf}

\IEEEoverridecommandlockouts %
\usepackage{amsmath} %
\usepackage{amssymb} %
\usepackage{amsfonts}
\usepackage{cite}
\usepackage[dvipsnames]{xcolor}
\usepackage[table]{xcolor}
\usepackage{graphicx}
\usepackage{tabularx}
\usepackage{multirow}
\usepackage{afterpage}
\usepackage{mathtools}
\usepackage{esvect} %
\usepackage[binary-units=true]{siunitx}
\usepackage{caption}
\usepackage[pdftex, pdfstartview={FitV},bookmarksopen=true,plainpages = false, colorlinks=true, linkcolor=black, citecolor = black, urlcolor = black,filecolor=black , pagebackref=false,hypertexnames=false, plainpages=false, pdfpagelabels ]{hyperref}
\usepackage[T1]{fontenc} %
\usepackage{mathtools, cuted} %

\usepackage{balance} %
\usepackage{booktabs} %
\usepackage[export]{adjustbox} %

\newcolumntype{x}[1]{%
>{\raggedleft\hspace{0pt}}p{#1}}%

\usepackage{algorithm}
\usepackage[noend]{algpseudocode} %
\definecolor{commentclr}{RGB}{34, 139, 34}

\usepackage{tikz}

\usepackage{subcaption}
\DeclareCaptionLabelSeparator{periodspace}{.\quad}
\usepackage{amsthm}

\theoremstyle{definition}

\usepackage{letltxmacro}
\LetLtxMacro\orgvdots\vdots
\LetLtxMacro\orgddots\ddots

\makeatletter
\DeclareRobustCommand\vdots{%
	\mathpalette\@vdots{}%
}
\newcommand*{\@vdots}[2]{%
	\sbox0{$#1\cdotp\cdotp\cdotp\m@th$}%
	\sbox2{$#1.\m@th$}%
	\vbox{%
		\dimen@=\wd0 %
		\advance\dimen@ -3\ht2 %
		\kern.5\dimen@
		\dimen@=\wd2 %
		\advance\dimen@ -\ht2 %
		\dimen2=\wd0 %
		\advance\dimen2 -\dimen@
		\vbox to \dimen2{%
			\offinterlineskip
			\copy2 \vfill\copy2 \vfill\copy2 %
		}%
	}%
}
\DeclareRobustCommand\ddots{%
	\mathinner{%
		\mathpalette\@ddots{}%
		\mkern\thinmuskip
	}%
}
\newcommand*{\@ddots}[2]{%
	\sbox0{$#1\cdotp\cdotp\cdotp\m@th$}%
	\sbox2{$#1.\m@th$}%
	\vbox{%
		\dimen@=\wd0 %
		\advance\dimen@ -3\ht2 %
		\kern.5\dimen@
		\dimen@=\wd2 %
		\advance\dimen@ -\ht2 %
		\dimen2=\wd0 %
		\advance\dimen2 -\dimen@
		\vbox to \dimen2{%
			\offinterlineskip
			\hbox{$#1\mathpunct{.}\m@th$}%
			\vfill
			\hbox{$#1\mathpunct{\kern\wd2}\mathpunct{.}\m@th$}%
			\vfill
			\hbox{$#1\mathpunct{\kern\wd2}\mathpunct{\kern\wd2}\mathpunct{.}\m@th$}%
		}%
	}%
}
\makeatother

\let\oldnl\nl%
\newcommand{\nonl}{\renewcommand{\nl}{\let\nl\oldnl}}%
\makeatother

\def\sl{\mathcal{L}}

\graphicspath{{figures/}}

\usepackage{tabularx}
\usepackage{array} %
\usepackage{makecell}
\usepackage{multirow}
\usepackage[acronym,nomain]{glossaries}
\glsdisablehyper %
\newacronym{uav}{UAV}{unmanned aerial vehicle}
\newacronym{gnss}{GNSS}{global navigation satellite system}
\newacronym{ins}{INS}{inertial navigation system}
\newacronym{avl}{AVL}{absolute visual localization}

\makeatletter
\def\ps@conferenceheader{%
  \ps@empty
  \def\@oddhead{%
    \raisebox{0pt}[0pt][0pt]{%
      \parbox[b]{\textwidth}{%
        \normalfont\sffamily\fontsize{8}{10}\selectfont\raggedright
        2026 IEEE/RSJ International Conference on Intelligent Robots and Systems (IROS)\\
        September 27 - October 1, 2026. Pittsburgh, PA, USA
      }%
    }%
  }%
  \let\@evenhead\@oddhead
  \def\@oddfoot{%
    \parbox[b]{\textwidth}{%
      \normalfont\fontsize{7}{8}\selectfont
      \textcopyright\ 2026 IEEE. Personal use of this material is permitted.
      Permission from IEEE must be obtained for all other uses, in any current
      or future media, including reprinting/republishing this material for
      advertising or promotional purposes, creating new collective works, for
      resale or redistribution to servers or lists, or reuse of any copyrighted
      component of this work in other works.%
    }%
  }%
  \let\@evenfoot\@oddfoot
}
\makeatother

\begin{document}

\title{
Large-Scale Geometric Map-Based Localization of UAVs in GNSS-Denied Urban Environments}

\author{Garth J.S. Terlizzi III and Kaveh Fathian%
 \thanks{The authors are with the Department of Computer Science, Colorado School of Mines; Emails: garth\_terlizzi@mines.edu, fathian@ariarobotics.com.}
 \thanks{The authors used a Large Language Model (OpenAI, GPT-5.2) for minor language editing in portions of the manuscript. All technical content and results were developed and verified by the authors, who take full responsibility.}
}%

\maketitle
\thispagestyle{conferenceheader}
\bstctlcite{IEEEexample:BSTcontrol}

\begin{abstract}
\Glspl{uav} operating in \glsentryshort{gnss}-denied urban
environments require alternative methods for position estimation. Existing approaches based on satellite image retrieval or learned descriptors are sensitive to appearance variation and degrade rapidly as the search area grows. We present a vision-based localization system that matches building patterns observed from a downward-facing UAV camera against a reference building footprint database. Our approach detects buildings in aerial imagery, accumulates observations across frames into a unified map, and matches local building arrangements against reference footprints using a novel geometry-driven descriptor that augments local triangle structure with per-building shape features. By encoding spatial relationships between nearby buildings rather than visual appearance, the system is robust to appearance variations and remains discriminative over large search areas.
Evaluations on seven flights across four municipalities in a large metropolitan area demonstrate $100\%$ Recall@1 at search areas of ${\approx}113$\,km$^2$ and $254$\,km$^2$, and $71.4\%$ Recall@1 when expanded to ${\approx}452$\,km$^2$, encompassing up to $277,000$ buildings.
In contrast, baseline methods degrade rapidly and achieve $0\%$ Recall@1 at $254$\,km$^2$ and $452$\,km$^2$.
\end{abstract}

\section{Introduction}\label{sec:intro}

Advances in onboard sensing, autonomy, and communication networks have transformed \glspl{uav} into strategically important aerial systems with growing deployment across defense, critical infrastructure monitoring, disaster response, and commercial logistics~\cite{ahmed_recent_2022}. Most autonomous \glspl{uav} rely on \glspl{gnss} and \glspl{ins} for positioning, but \gls{gnss} signals are highly vulnerable to jamming, spoofing, and signal dropout in urban canyons or contested environments~\cite{couturier_review_2021}. This vulnerability has motivated a broad class of problems in \gls{avl}, wherein a \gls{uav} must estimate its global pose using only onboard sensing and a pre-existing reference map, without reliance on \gls{gnss}~\cite{couturier_review_2024}.

\begin{figure}[t!]
\centering
\includegraphics[trim = 10mm 20mm 0mm 60mm, width=1.\columnwidth]{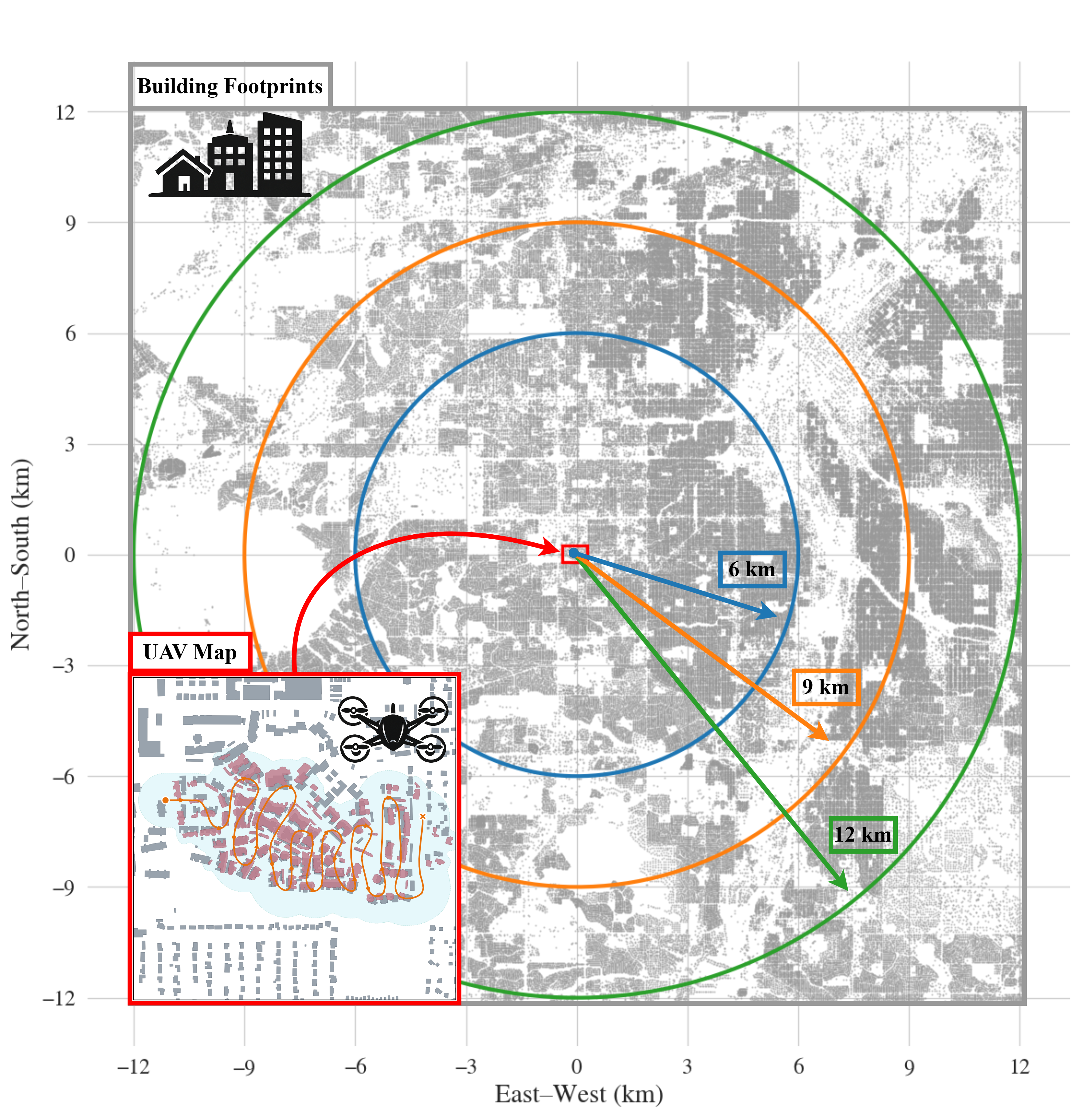}
\caption{Successful localization example using our proposed approach, across areas spanning ${\approx}113$\,km$^2$ (blue circle), ${\approx}254$\,km$^2$ (orange circle), and ${\approx}452$\,km$^2$ (green circle). Our novel star descriptors, generated from observed buildings, are matched across the \gls{uav} and reference maps for localization. The \gls{uav} map contains only $104$ observed buildings, yet is successfully matched against up to $272{,}878$ reference buildings, demonstrating the large-scale capability of our method.}
\label{fig:opening}
\end{figure}

\begin{figure*}[t]
\centering
\includegraphics[trim = 0mm 0mm 0mm 0mm, clip, width=0.99\textwidth] {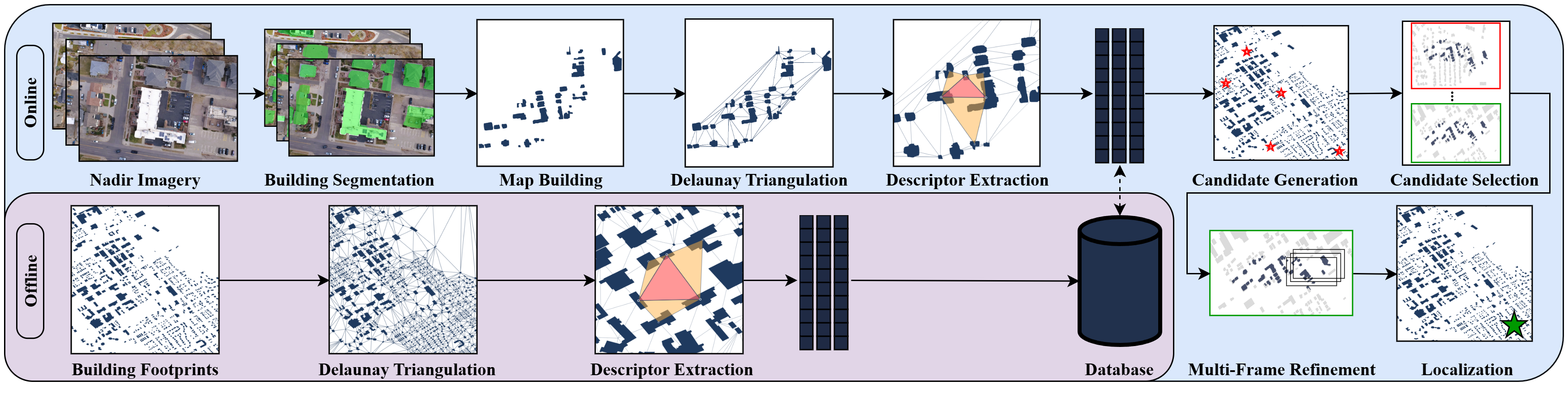}
\caption{Overview of the proposed localization pipeline. Aerial keyframes are segmented for buildings, stitched into an internal map, triangulated, and matched against a reference building footprint database via star descriptors and consensus voting.}
\vspace*{-0.5em}
\label{fig:pipeline}
\end{figure*}

However, traditional \gls{avl} methods face significant challenges when matching \gls{uav} imagery to reference maps, including scale and rotation ambiguity~\cite{ye_exploring_2025}, seasonal appearance discrepancies~\cite{kinnari_season-invariant_2022, fragoso_seasonally_2021}, and significant perspective changes~\cite{chen_oblique-robust_2024}. To address these issues, this paper introduces a localization system specifically designed for urban and suburban environments that leverages geometric building pattern matching. The system detects building patterns in downward-facing drone imagery and matches them against reference building footprints from the Microsoft Building Footprints database~\cite{heris_rasterized_2020}.

Our core contribution is a localization system that remains discriminative over search areas far beyond those at which appearance-based methods remain reliable, by leveraging Delaunay-based star descriptors~\cite{li_localization_2020} augmented with per-building shape features. Candidate locations are generated by matching these descriptors against a reference database and ranked through consensus voting and mutual nearest-neighbor scoring, with a final multi-frame refinement stage that corrects residual stitching drift.

\textbf{Contributions:}
The main contributions of this work are:
\begin{itemize}
    \item A complete pipeline for \gls{gnss}-denied \gls{uav} localization using downward-facing monocular imagery, lightweight platform telemetry (altitude and gimbal attitude), and a publicly available building footprint database. Our geometry-first reference design avoids satellite texture matching, reducing sensitivity to reference appearance shifts and orthorectification artifacts.

    \item A novel building-geometry descriptor that augments Delaunay triangle structure---which alone becomes ambiguous in dense urban scenes---with per-building shape attributes (compactness, elongation, rectangularity, convexity), extending prior triangle-only formulations~\cite{li_localization_2020} to retain discriminability at metropolitan scale.
     
    \item A scalable matching strategy, evaluated on real-world drone flight data and achieving successful localizations in all $7$ experiments at search radii of $6$\,km (${\approx}113$\,km$^2$) and $9$\,km (${\approx}254$\,km$^2$), and $5/7$ at $12$\,km (${\approx}452$\,km$^2$), outperforming existing baseline methods at large areas.

\end{itemize}

\section{Related Work}\label{sec:related}

\gls{avl} methods can be organized along three complementary design axes: (i) temporal evidence usage, (ii) reference map representation, and (iii) architecture type.

\textbf{Single-frame vs. multi-frame localization.} Localization strategies differ in whether they estimate pose from an isolated observation or accumulate evidence across a sequence. Single-frame methods typically cast \gls{avl} as image retrieval or Visual Place Recognition (VPR), matching a query \gls{uav} image to the most similar georeferenced reference image to obtain a coarse global estimate~\cite{he_leveraging_2024}. These methods are computationally efficient and effective for initialization, but can be ambiguous in visually repetitive or weakly structured environments, where multiple candidate locations may exhibit similar appearance~\cite{kinnari_season-invariant_2022, kinnari_lsvl_2023}. Multi-frame methods reduce this ambiguity by exploiting temporal continuity along the \gls{uav} trajectory~\cite{wang_sequence_2024}. Prior work incorporates sequential evidence using probabilistic filtering (e.g., Monte Carlo localization)~\cite{kinnari_season-invariant_2022, mantelli_novel_2019}, optimization-based fusion such as pose-graph optimization~\cite{zhang_georvlf_2025}, or sequence-consistency constraints and trajectory-aligned matching~\cite{kinnari_lsvl_2023, goforth_gps-denied_2019}. By integrating relative motion and repeated observations, these methods improve robustness and stabilize global pose estimates.

\textbf{Reference map representation.} The reference map modality strongly shapes both sensing assumptions and downstream matching. Satellite imagery remains dominant in \gls{uav} \gls{avl} because it offers dense georeferenced coverage and enables direct visual comparison with aerial queries~\cite{couturier_review_2024}. Some pipelines also pair satellite imagery with topography-derived map cues for improved cross-view matching~\cite{chen_real-time_2021}. However, satellite pipelines must handle strong appearance gaps between \gls{uav} and map views, seasonal and illumination variation, and orthorectification artifacts~\cite{kinnari_season-invariant_2022, fragoso_seasonally_2021, chen_oblique-robust_2024}. Vector and semantic map representations instead encode cartographic structure (e.g., buildings, roads, land-use regions), yielding compact and appearance-invariant references~\cite{wang_vecmaplocnet_2025}. Prior work has explored building-ratio maps~\cite{choi_brm_2020} and map-like translated representations~\cite{schleiss_translating_2019,wang_vecmaplocnet_2025}, often improving robustness to seasonal change at the cost of reduced texture detail and vulnerability to map incompleteness.

\textbf{Localization architecture.} Recent work can be grouped by matching strategy. \textit{Retrieval-based models} (e.g., NetVLAD adaptations~\cite{arandjelovic_netvlad_2018}) perform large-scale database search using learned global embeddings for coarse localization~\cite{xiao_long-range_2023}, often followed by local geometric matching~\cite{he_leveraging_2024, li_jointly_2023, chen_real-time_2021}, sequence constraints~\cite{wang_sequence_2024, meng_airgeonet_2024}, or pose-graph/odometry fusion~\cite{zhang_georvlf_2025} to improve final pose precision. \textit{Pairwise cross-view matching models} learn direct similarity between \gls{uav} observations and candidate map patches, often with metric-learning objectives for robustness under viewpoint and appearance shifts~\cite{kinnari_season-invariant_2022, lin_gnss-denied_2025,sun_f3-net_2023,dai_transformer-based_2022}. 

In parallel, there is emerging interest in \textit{graph-based} localization architectures that encode relational structure among landmarks and match at graph level rather than only descriptor level~\cite{duan_scene_2024}. Recent Delaunay-based approaches by Winterton et al.~\cite{winterton_aerial_2025} and Bennett et al.~\cite{bennett_bag--graph-attributes_2025} show promising graph-based scalability (Bennett reports up to 822.2~km$^2$), though both are evaluated on high-altitude orthorectified imagery rather than real low-altitude drone footage. Among methods evaluated on real drone footage, the largest reported reference area to our knowledge is $100$\,km$^2$ by Kinnari et al.\ (LSVL)~\cite{kinnari_lsvl_2023}, which motivates our evaluation at ${\approx}254$\,km$^2$. Our method is multi-frame, operates on vector map representations, and employs a graph-based architecture using geometric star descriptors derived from Delaunay triangulations.

\section{Proposed Pipeline}\label{sec:pipeline}

Fig.~\ref{fig:pipeline} shows the components of the proposed pipeline, as explained in detail below.
We first summarize offline reference preparation, then describe the online stages in order.

\subsection{Offline Reference Preparation}

Before deployment, Microsoft Building Footprints~\cite{heris_rasterized_2020} within a predetermined search radius are processed offline to produce the database used for online localization. The offline workflow mirrors the online geometric pipeline: building centroids are extracted, a Delaunay triangulation is computed, and star descriptors are formed to capture local triangle geometry and neighborhood context. These descriptors are stored in an indexed database that enables fast candidate retrieval during online matching. The triangulation and descriptor definitions follow the same formulation used online (see Sections~\ref{sec:delaunay} and~\ref{sec:descriptor}); candidate retrieval is described in Section~\ref{sec:candidate-generation}.
\subsection{Nadir Imagery and Building Segmentation}

The proposed system operates using a lightweight onboard sensor suite consisting of a downward-facing monocular RGB camera, altitude telemetry, and gimbal attitude telemetry. Each incoming image frame is processed by a semantic segmentation network to identify building pixels. We employ a SegFormer-B5 transformer-based segmentation model~\cite{xie_segformer_2021} trained on the INRIA Aerial Image Labeling Dataset~\cite{maggiori_can_2017}, configured for binary building--background classification. The output of this stage is a per-frame binary building mask; logits are upsampled to image size and thresholded to produce the final mask. A keyframe selection policy retains only keyframes where the camera in-plane rotation has changed by more than $15^\circ$ since the last keyframe or at least $5$\,s have elapsed, whichever condition is met first.
\begin{figure}[!t]
\centering
\begin{subfigure}[t]{0.48\columnwidth}
    \centering
    \includegraphics[width=\linewidth]{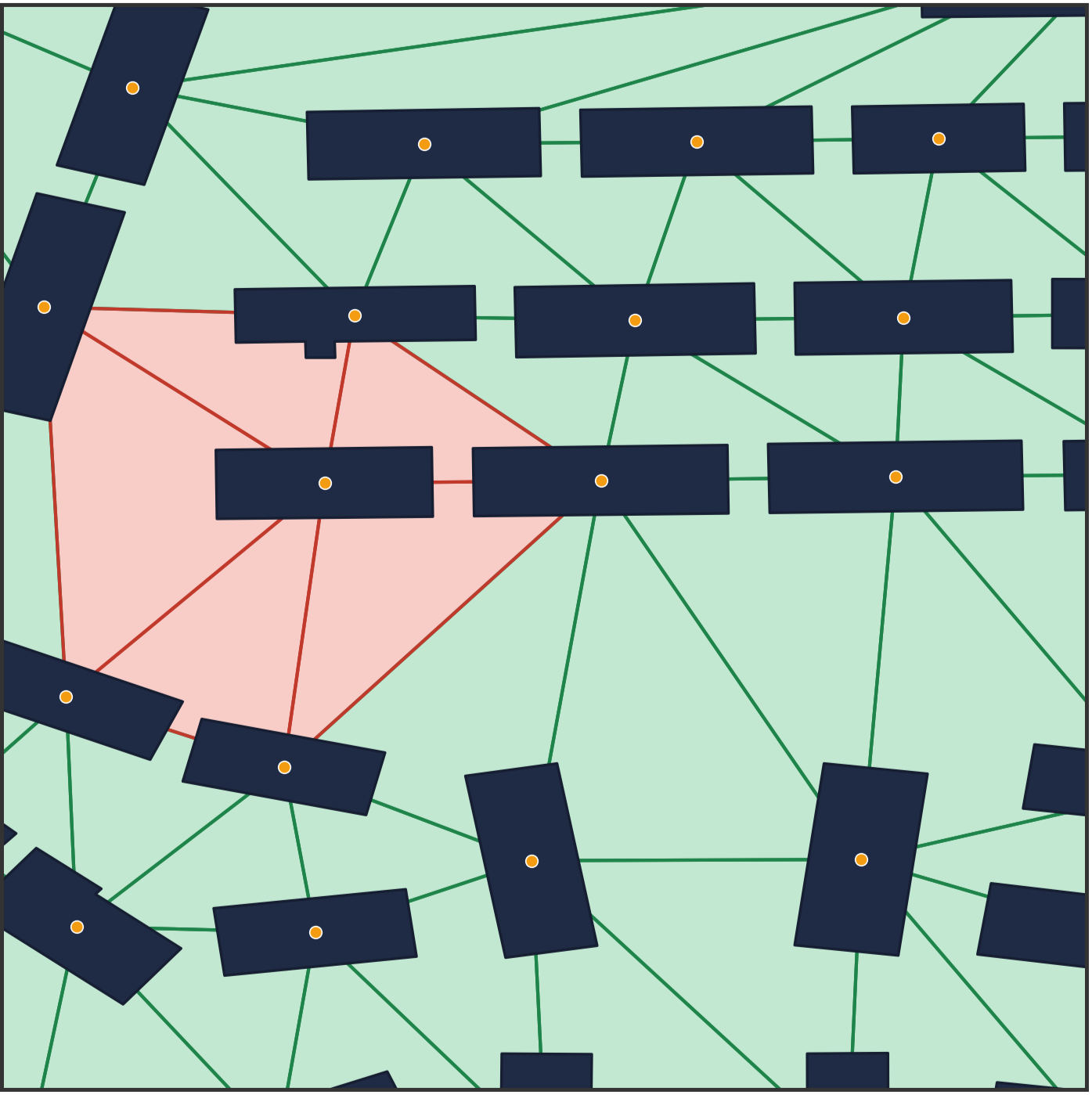}
    \caption{Building Footprints}
    \label{fig:delaunay_footprints}
\end{subfigure}
\hfill
\begin{subfigure}[t]{0.48\columnwidth}
    \centering
    \includegraphics[width=\linewidth]{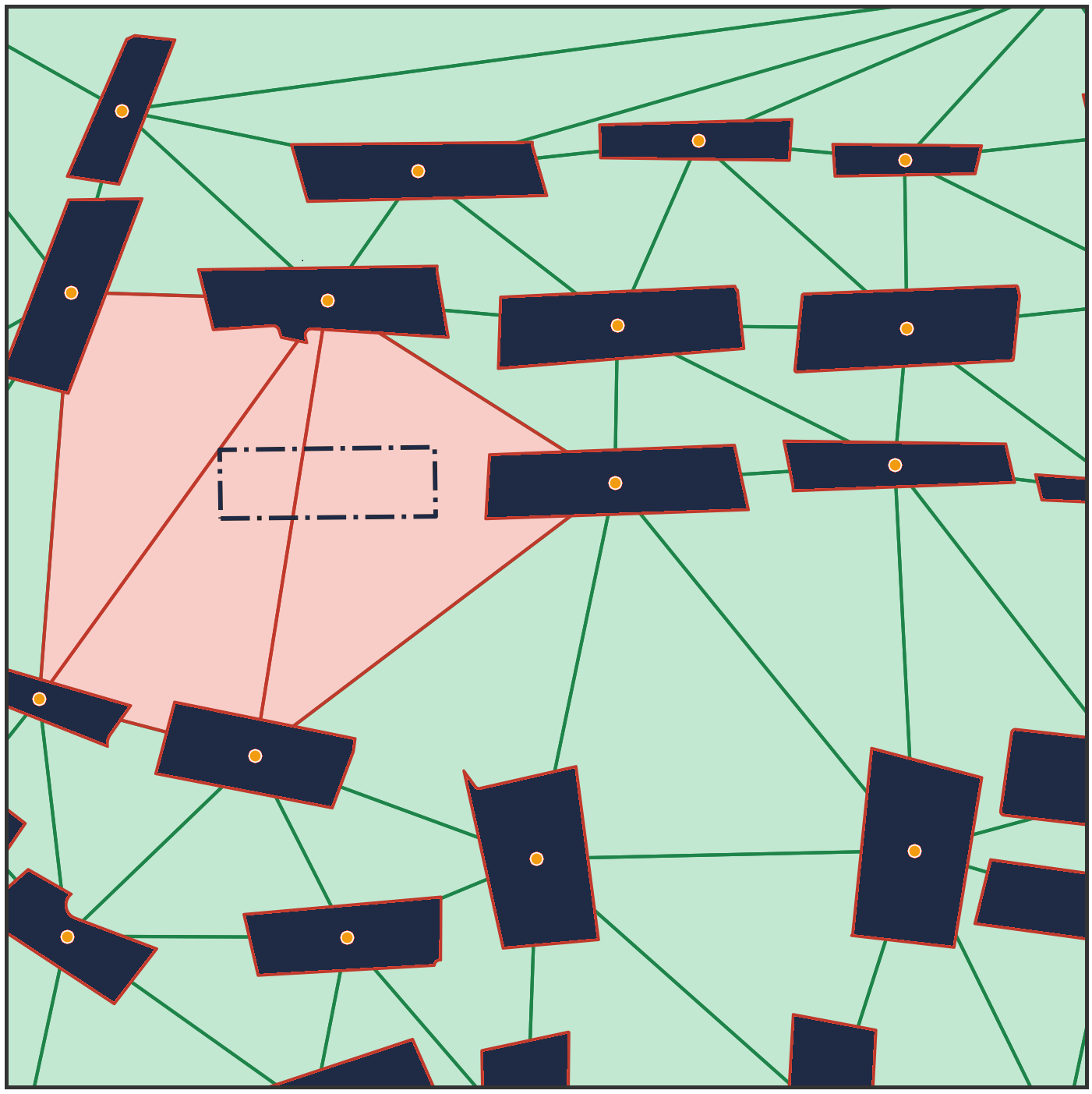}
    \caption{\gls{uav}-segmented Map}
    \label{fig:delaunay_segmented}
\end{subfigure}
\caption{Effect of a missed detection on the Delaunay triangulation. Left: ground truth from Microsoft Building Footprints. Right: \gls{uav}-segmented map with minor error (jitter and distortion) and a missed detection. Red triangles are significantly altered by the missed detection, while green triangles retain their shape, highlighting Delaunay's neighborhood-preservation behavior under local segmentation noise.}
\label{fig:delaunay_effect}
\end{figure}

\subsection{Map Building} 

To reduce single-keyframe noise and partial observability, we fuse building observations across consecutive keyframes into a local internal map before global matching. Adjacent keyframes are aligned by extracting LoFTR~\cite{sun_loftr_2021} correspondences and estimating a planar similarity transform (rotation, translation, uniform scale) with RANSAC~\cite{fischler_random_1981}. We use similarity transforms rather than full homographies because nadir-view geometry and altitude telemetry constrain perspective effects and improve stability. Keyframe   
  pairs are rejected if RANSAC yields too few inliers or if the estimated scale deviates significantly from unity, indicating a degenerate fit. Accepted transforms are then scale-corrected using the altitude ratio between keyframes to prevent drift from minor altitude variation. 

Aligned masks are then fused using observation-count filtering: each pixel's building classification is retained only if it is confirmed across multiple overlapping keyframes, suppressing transient segmentation artifacts.

To mitigate odometry drift, we optimize a pose graph with per-keyframe state \([x,y,\theta,\log s]\), where \(x,y\) are the keyframe-center pixel coordinates, \(\theta\) is the keyframe rotation angle in radians, and \(\log s\) is the log of the cumulative scale factor. Edges include consecutive relative poses from feature matching and loop-closure relative poses from spatially overlapping revisits. Loop closures are identified by testing spatially proximate prior keyframes via LoFTR matching. We optimize the pose graph using a robust nonlinear least-squares formulation (Huber loss) via the GTSAM library~\cite{dellaert_factor_2012}, incorporating inlier-weighted residuals, increased weighting on loop-closure constraints, and a unit-scale prior. The optimized keyframe poses are used to re-composite the per-keyframe segmentation masks into a drift-corrected internal map, from which building centroids and shape descriptors are extracted via connected-component analysis.

\begin{table}[!t]
\scriptsize
\centering
\caption{Star descriptor features. Each triangle (central and three neighbors) is described by eight features: sorted interior angles and perimeter from triangle geometry, plus four building-shape metrics averaged over the three vertex buildings.}
\label{tab:descriptor}
\begin{tabular*}{\linewidth}{@{\extracolsep{\fill}}llc@{}}
\toprule
\textbf{Feature} & \textbf{Definition} & \textbf{Dim} \\
\midrule
\rowcolor{gray!19}
\multicolumn{3}{@{}l}{\textit{Triangle geometry}} \\
Sorted angles        & $\alpha_1 \leq \alpha_2 \leq \alpha_3$
               & $3$ \\
Perimeter            & $P = \sum \|e_i\|$
               & $1$ \\
\midrule
\rowcolor{gray!15}
\multicolumn{3}{@{}l}{\textit{Building shape (avg.)}} \\
Compactness          & $4\pi A \,/\, L^2$
               & $1$ \\
Elongation           & $w^{\min} / w^{\max}$
               & $1$ \\
Rectangularity       & $A \,/\, (w^{\min} w^{\max})$
               & $1$ \\
Convexity            & $A \,/\, A_{\text{hull}}$
               & $1$ \\
\midrule
\multicolumn{2}{@{}l}{\textit{Per-triangle subtotal}} & $8$ \\
$1$ central $+$ $3$ neighbors & & \textbf{$32$} \\
\bottomrule
\end{tabular*} \\

\vspace{0.2em}
\begin{minipage}{\linewidth}
\scriptsize
$\alpha_i$: Interior angles; $e_i$: Triangle edge lengths; $A$: Building footprint area; $L$: Building perimeter; $w^{\min}$, $w^{\max}$: Short and long sides of the minimum rotated bounding rectangle; $A_{\text{hull}}$: Convex hull area.
\end{minipage}
\end{table}

\subsection{Delaunay Triangulation}\label{sec:delaunay}

Let \(\mathcal{P}=\{p_i\}_{i=1}^{N}\), \(p_i\in\mathbb{R}^2\), denote the building centroids in the stitched internal map. We compute the planar Delaunay triangulation \(\mathcal{T}_D(\mathcal{P})\)~\cite{barber_quickhull_1996}, where each triangle satisfies the empty-circumcircle condition (its circumcircle contains no other point in \(\mathcal{P}\)).
An example comparing triangulations from reference and segmented buildings is shown in Fig.~\ref{fig:delaunay_effect}.

The Delaunay triangulation induces a geometry-driven adjacency graph over $\mathcal{P}$, where two centroids are adjacent if they share an edge in $\mathcal{T}_D(\mathcal{P})$. This construction depends only on point geometry and avoids user-defined neighborhood thresholds (e.g., fixed radius or $k$-nearest neighbors). By maximizing the minimum interior angle among all triangulations of $\mathcal{P}$, it reduces poorly conditioned triangles and improves robustness to small centroid perturbations.

The Delaunay triangulation is equivariant under similarity transforms: for \(g(x)=sRx+t\) (\(s>0\), \(R\in \mathrm{SO}(2)\), \(t\in\mathbb{R}^2\)) and non-degenerate point sets, \(\mathcal{T}_D(g(\mathcal{P})) = g(\mathcal{T}_D(\mathcal{P}))\). In particular, the combinatorial structure---which triples of centroids form triangles and which triangles share edges---is invariant under such transforms. Therefore, global translation, rotation, and uniform scale do not change local triangle topology.

\subsection{Descriptor Extraction}\label{sec:descriptor}
\begin{figure}[!t]
    \centering
{\setlength{\fboxsep}{2pt}\setlength{\fboxrule}{0.4pt}\fbox{\includegraphics[width=.99\linewidth]{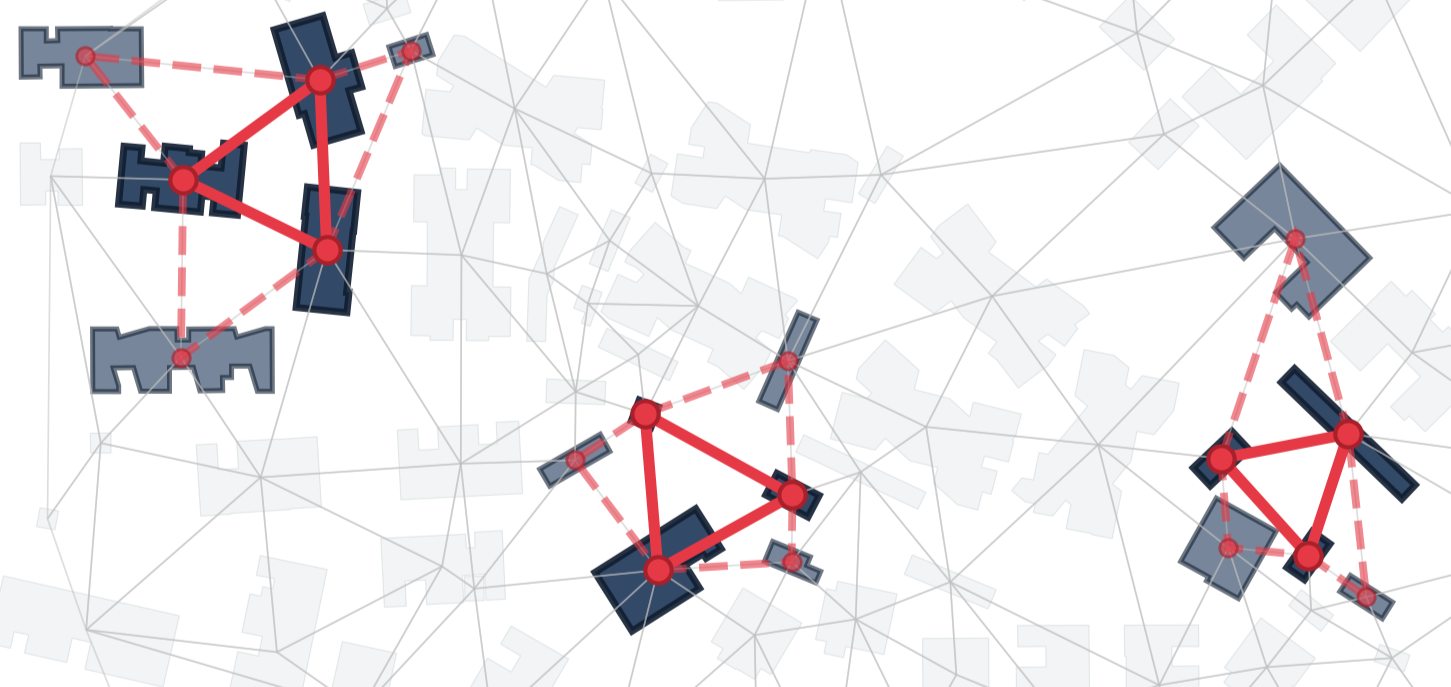}}}
    \caption{Example star descriptors. The central triangle (consisting of the dark buildings) and its adjacent neighbors form a local geometric pattern that encodes both spatial arrangement and per-building shape features.}
    \label{fig:star_descriptors}
\end{figure}

To match local observations against large reference maps, we represent each region using a \textbf{star descriptor}, defined as a central Delaunay triangle and its three adjacent triangles following Li et al.~\cite{li_localization_2020}. We extend their star structure to the aerial domain, replacing their per-triangle geometry with sorted interior angles and perimeter, and augmenting each triangle's representation with node-level shape features for each vertex building. Li et al.\ applied this structure to ground-robot forest localization, matching a single lidar scan of tree-trunk landmarks against a local subgraph of a prior tree map, where point-like stems carry no shape and triangle geometry alone suffices. Buildings instead have distinctive footprints, and in dense urban scenes many triangles are near-congruent; the per-building shape features restore discriminability that triangle topology alone cannot provide.

For each triangle, we compute two groups of rotation-invariant features (Table~\ref{tab:descriptor}). The first group captures triangle geometry: sorted interior angles and perimeter. The second group captures building shape by averaging four polygon metrics over the three vertex buildings: compactness, elongation, rectangularity, and convexity. Together these yield eight features per triangle.

Each neighboring triangle is described by the same eight features. The three neighbors are ordered by distance from the central triangle's centroid. The star descriptor concatenates the central triangle's features with those of its three neighbors, producing a $32$-dimensional vector. Only triangles with exactly three neighbors form valid stars; boundary triangles are discarded. Edge lengths and perimeter are expressed in metric units via the altitude-derived ground sampling distance, ensuring comparability with reference footprint measurements.

\begin{table}[!t]
\scriptsize
\centering
\caption{Summary of \gls{uav} flight trajectories used for evaluation. In our environment labels, suburban refers to single-family homes, while residential refers to multi-family homes and apartment complexes.}
\label{tab:flight_summary}

\begin{tabular*}{\linewidth}{@{\extracolsep{\fill}}cccccc@{}}
\toprule
\textbf{Flight \#}
& \textbf{Length (km)}
& \textbf{Alt. (m)}
& \textbf{Keyframes}
& \makecell{\textbf{\# Buildings} \\ \textbf{Detected}}
& \textbf{Environment} \\
\midrule

$1$ & $0.39$ & $108$ & $20$ & $30$ & C \\
$2$ & $1.68$ & $116$ & $67$ & $78$ & C+S \\
$3$ & $2.19$ & $105$ & $101$ & $133$ & S \\
$4$ & $3.18$ & $115$ & $256$ & $104$ & R \\
$5$ & $2.26$ & $98$ & $76$ & $87$ & C+S \\
$6$ & $2.85$ & $106$ & $181$ & $121$ & C+S \\
$7$ & $2.02$ & $97$ & $186$ & $104$ & C+S+R \\

\bottomrule
\end{tabular*}

\vspace{0.2em}
\begin{minipage}{\linewidth}
\scriptsize
C: Commercial; S: Suburban; R: Residential
\end{minipage}
\end{table}

\subsection{Candidate Generation}\label{sec:candidate-generation}

Candidate matches are generated by matching every valid internal-map star descriptor against a large database of footprint-derived reference descriptors, using a two-stage procedure. A hard compatibility filter first rejects geometrically inconsistent pairs: writing $\alpha,\alpha'$, $P,P'$, and $\eta,\eta'$ for the central-triangle sorted angles, perimeters, and mean shape features (compactness, elongation, rectangularity, convexity) of a query and reference star, we define
\begin{equation}\label{eq:cg-terms}
d_\alpha=\lVert\alpha-\alpha'\rVert_1,\quad
d_P=\frac{\lvert P-P'\rvert}{\max(P,P')},\quad
d_S=\tfrac14\lVert\eta-\eta'\rVert_1,
\end{equation}
and reject a pair when $d_\alpha\ge\tau_\alpha$ or when $d_P$ or $d_S$ exceeds its tolerance ($\tau_\alpha=15^\circ$, $\tau_P=\tau_S=0.3$). Surviving pairs are then scored by a ratio-based similarity that weights geometry above shape ($2{:}1{:}1$, angle:perimeter:shape). Each term contributes a $[0,1]$ agreement suited to its type---angle to its $\tau_\alpha$ tolerance, perimeter as a scale-free ratio, shape directly:
\begin{equation}\label{eq:cg-score}
\sigma_\alpha{=}\max(0,\,1-d_\alpha/\tau_\alpha),\ \ \sigma_P{=}\frac{\min(P,P')}{\max(P,P')},\ \ \sigma_S{=}1-d_S,
\end{equation}
Per triangle these combine as $\tfrac14(2\sigma_\alpha+\sigma_P+\sigma_S)$, and the star score $s$ averages this over the four triangles, which are matched identically, so $s$ spans all $32$ descriptor dimensions. Only the top-ranked matches are retained.

Each match yields a candidate pose by estimating a $2$D similarity transform between the matched triangles, with vertex correspondence from the sorted interior-angle ordering: scale from matched edge-length ratios, rotation from the orientation offset of a matched edge pair, and translation from centroid alignment, mapping the query star into world coordinates as a candidate global \gls{uav} pose.

\begin{figure*}[!t]
    \centering

    \begin{subfigure}[t]{0.14\textwidth}
        \centering
        \includegraphics[width=\linewidth]{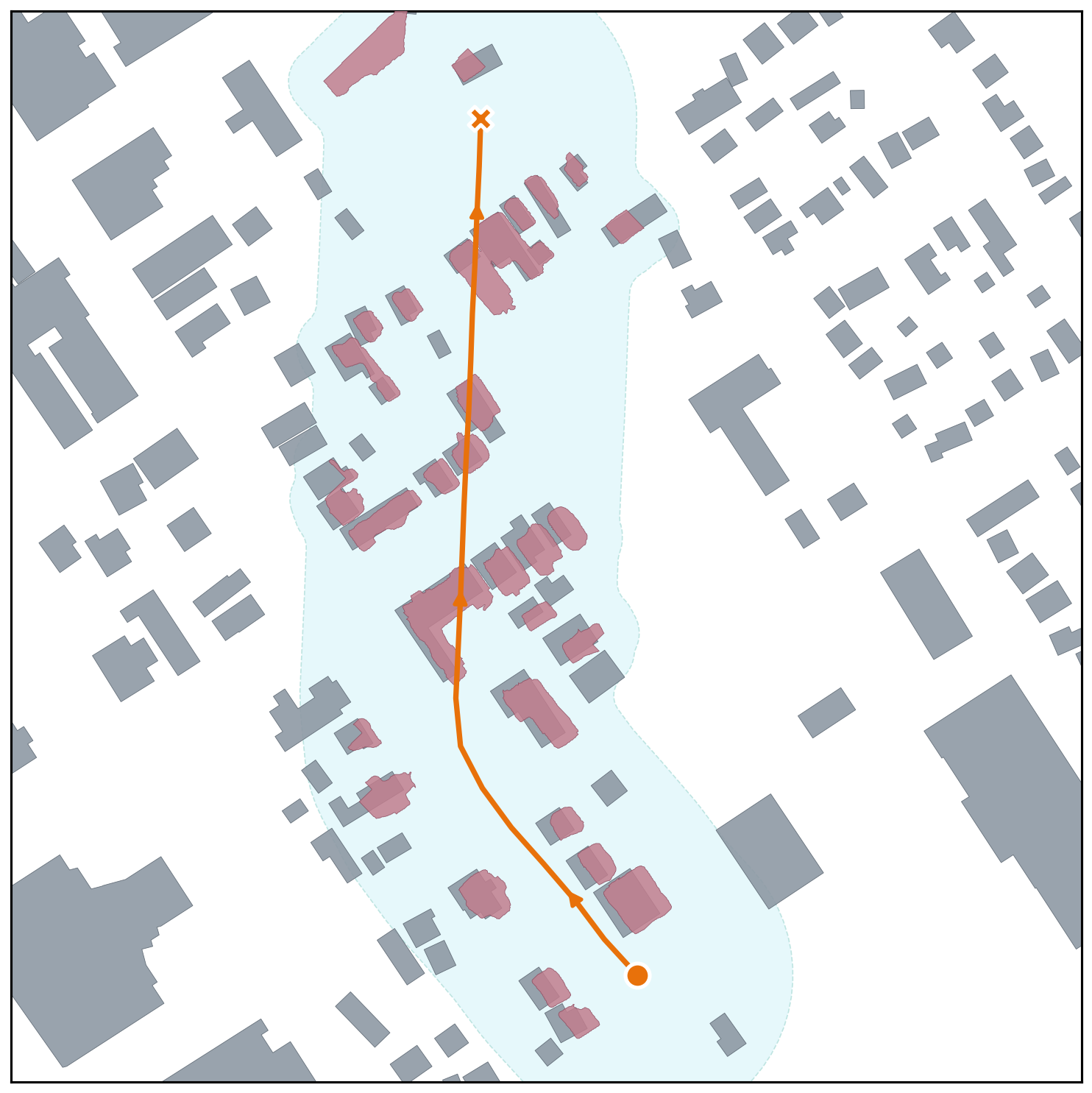}
        \caption{}
        \label{fig:panel_a}
    \end{subfigure}\hfill
    \begin{subfigure}[t]{0.14\textwidth}
        \centering
        \includegraphics[width=\linewidth]{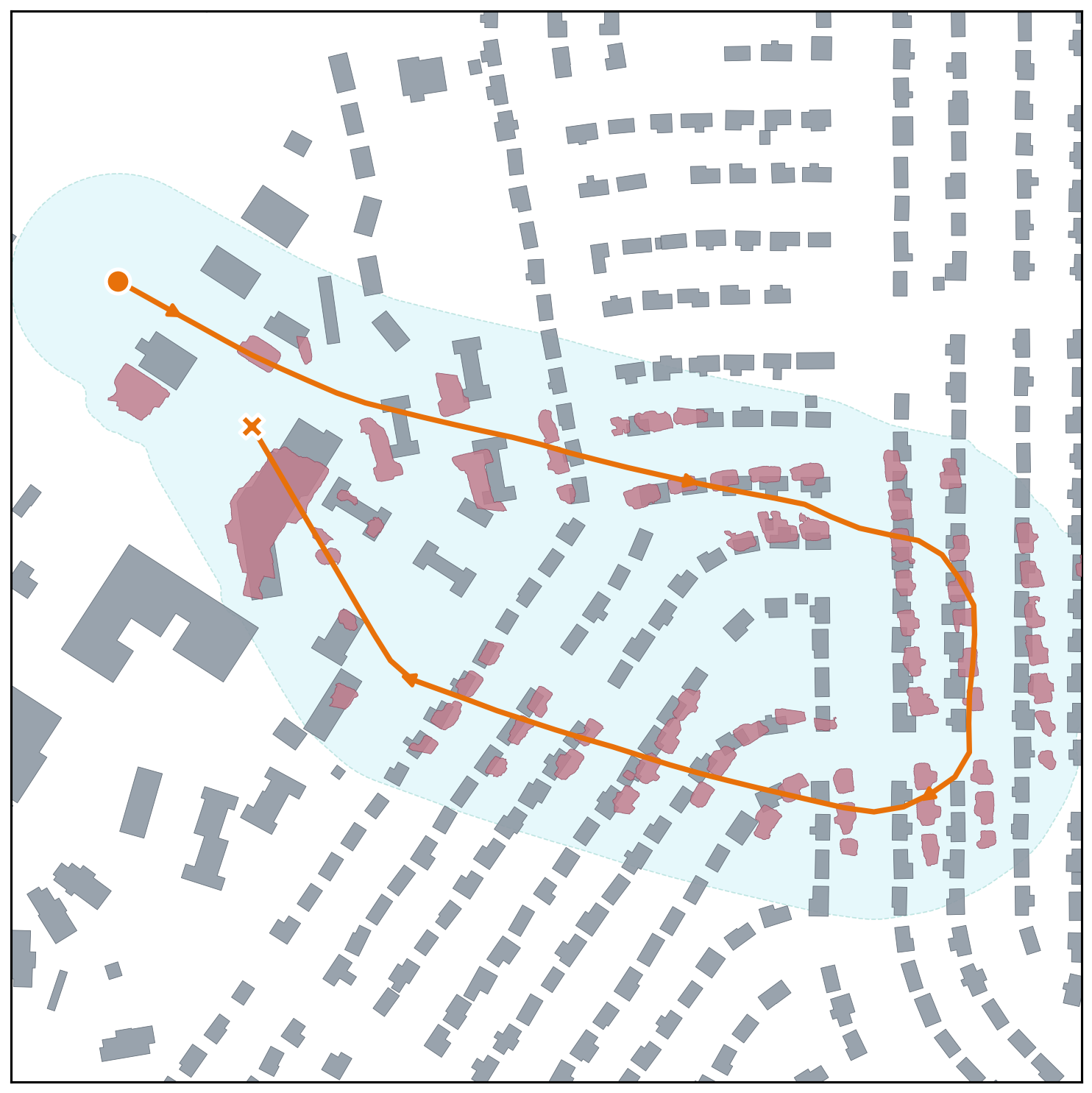}
        \caption{}
        \label{fig:panel_b}
    \end{subfigure}\hfill
    \begin{subfigure}[t]{0.14\textwidth}
        \centering
        \includegraphics[width=\linewidth]{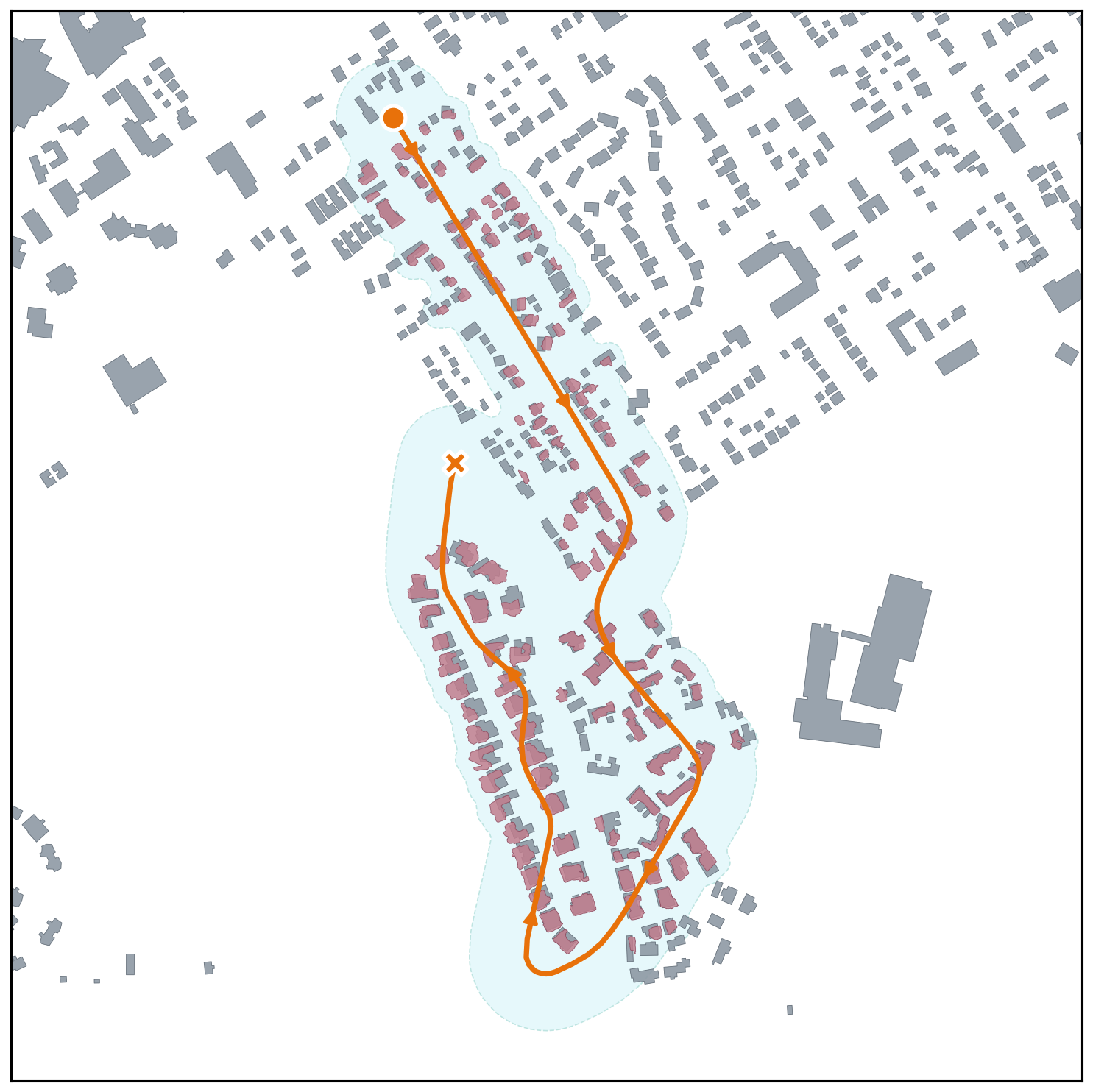}
        \caption{}
        \label{fig:panel_c}
    \end{subfigure}\hfill
    \begin{subfigure}[t]{0.14\textwidth}
        \centering
        \includegraphics[width=\linewidth]{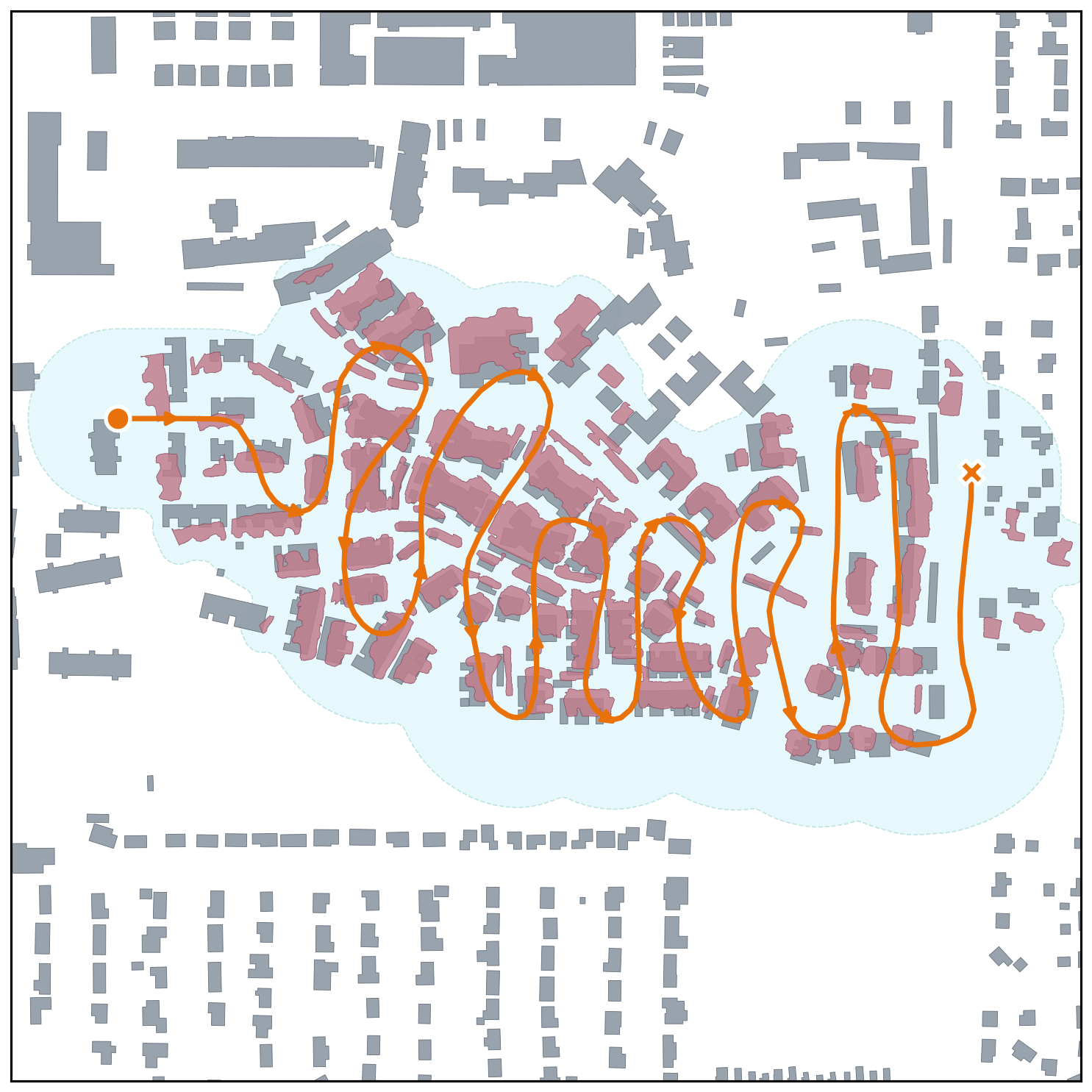}
        \caption{}
        \label{fig:panel_d}
    \end{subfigure}\hfill
    \begin{subfigure}[t]{0.14\textwidth}
        \centering
        \includegraphics[width=\linewidth]{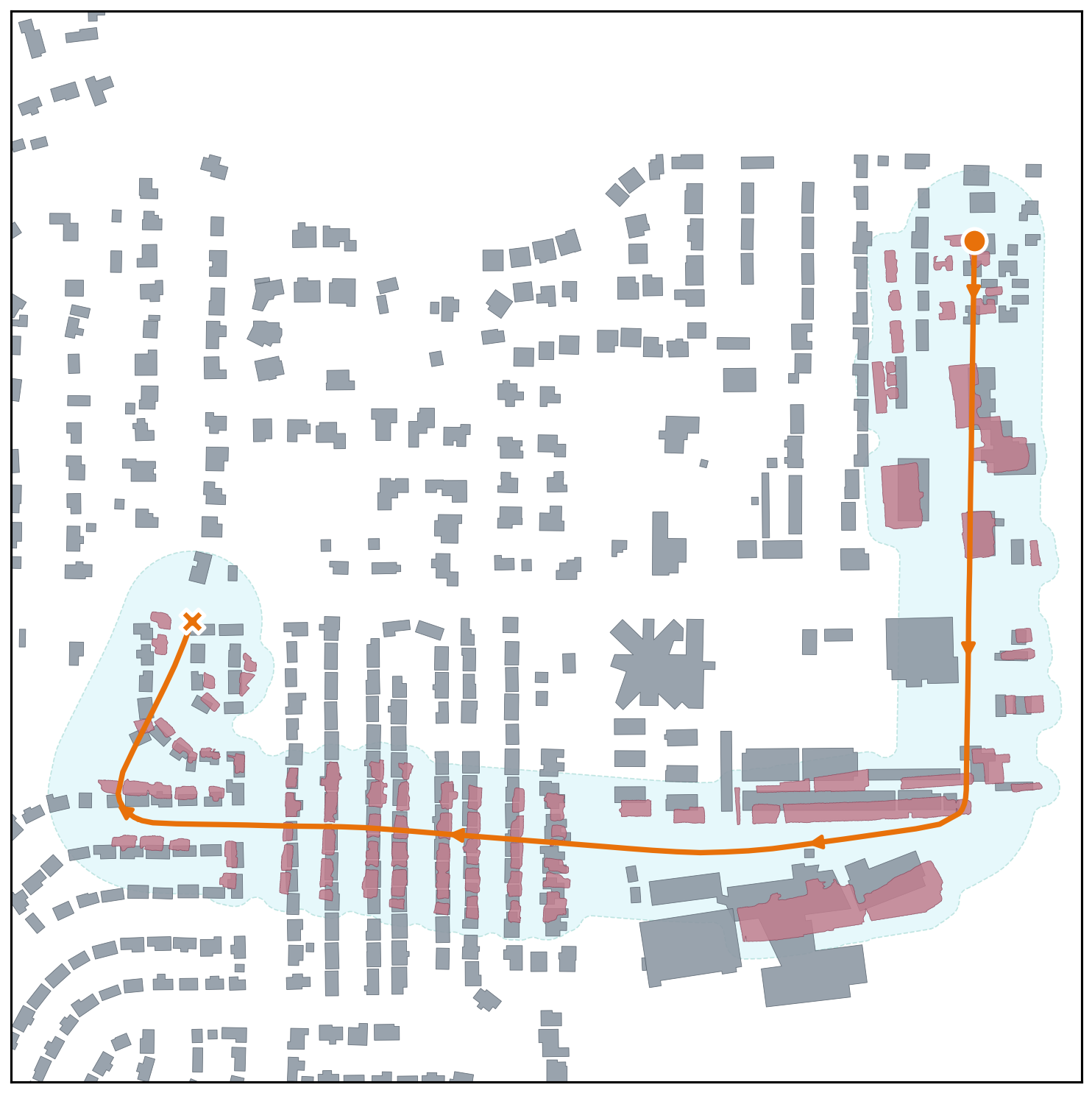}
        \caption{}
        \label{fig:panel_e}
    \end{subfigure}\hfill
    \begin{subfigure}[t]{0.14\textwidth}
        \centering
        \includegraphics[width=\linewidth]{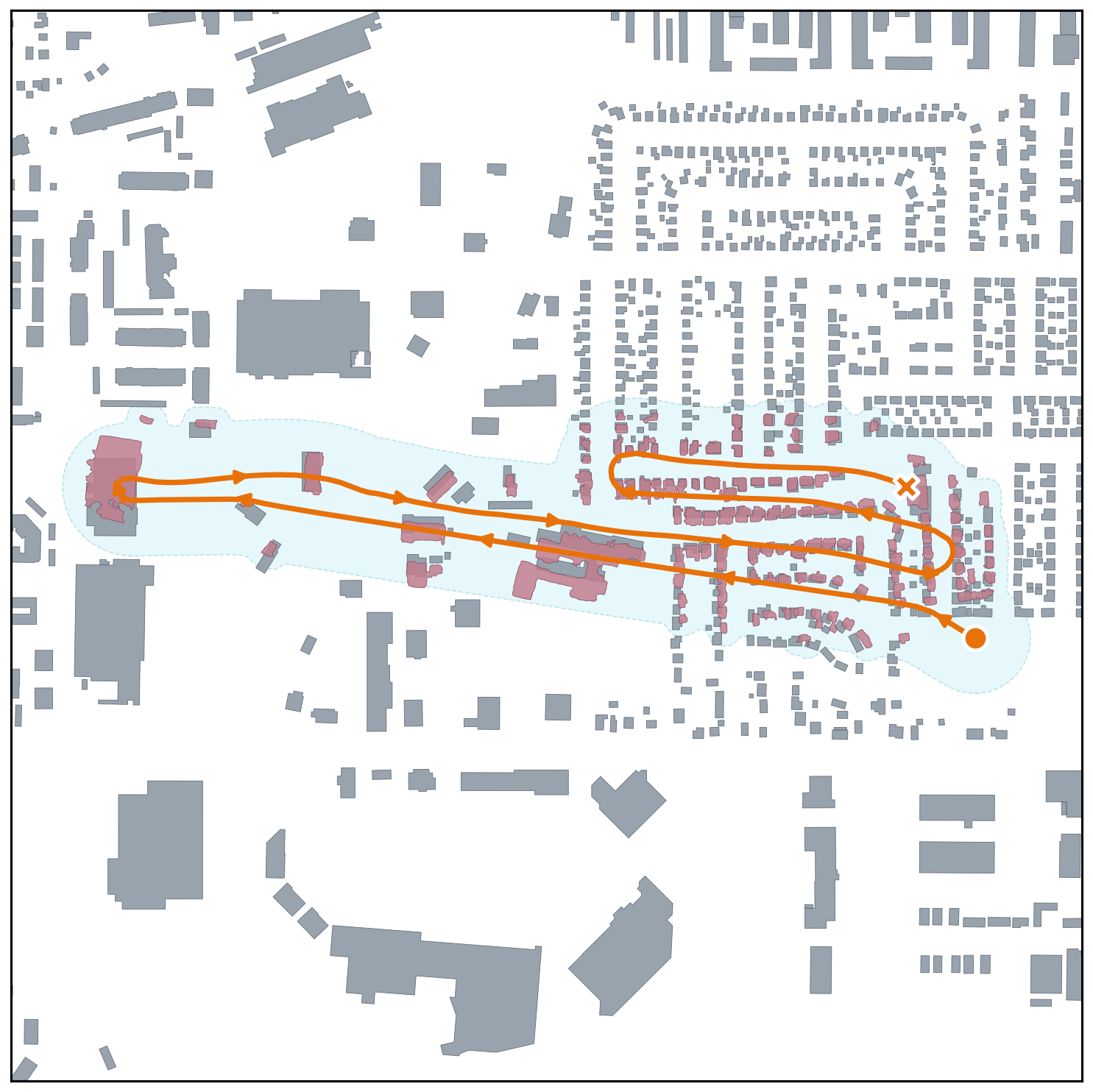}
        \caption{}
        \label{fig:panel_f}
    \end{subfigure}\hfill
    \begin{subfigure}[t]{0.14\textwidth}
        \centering
        \includegraphics[width=\linewidth]{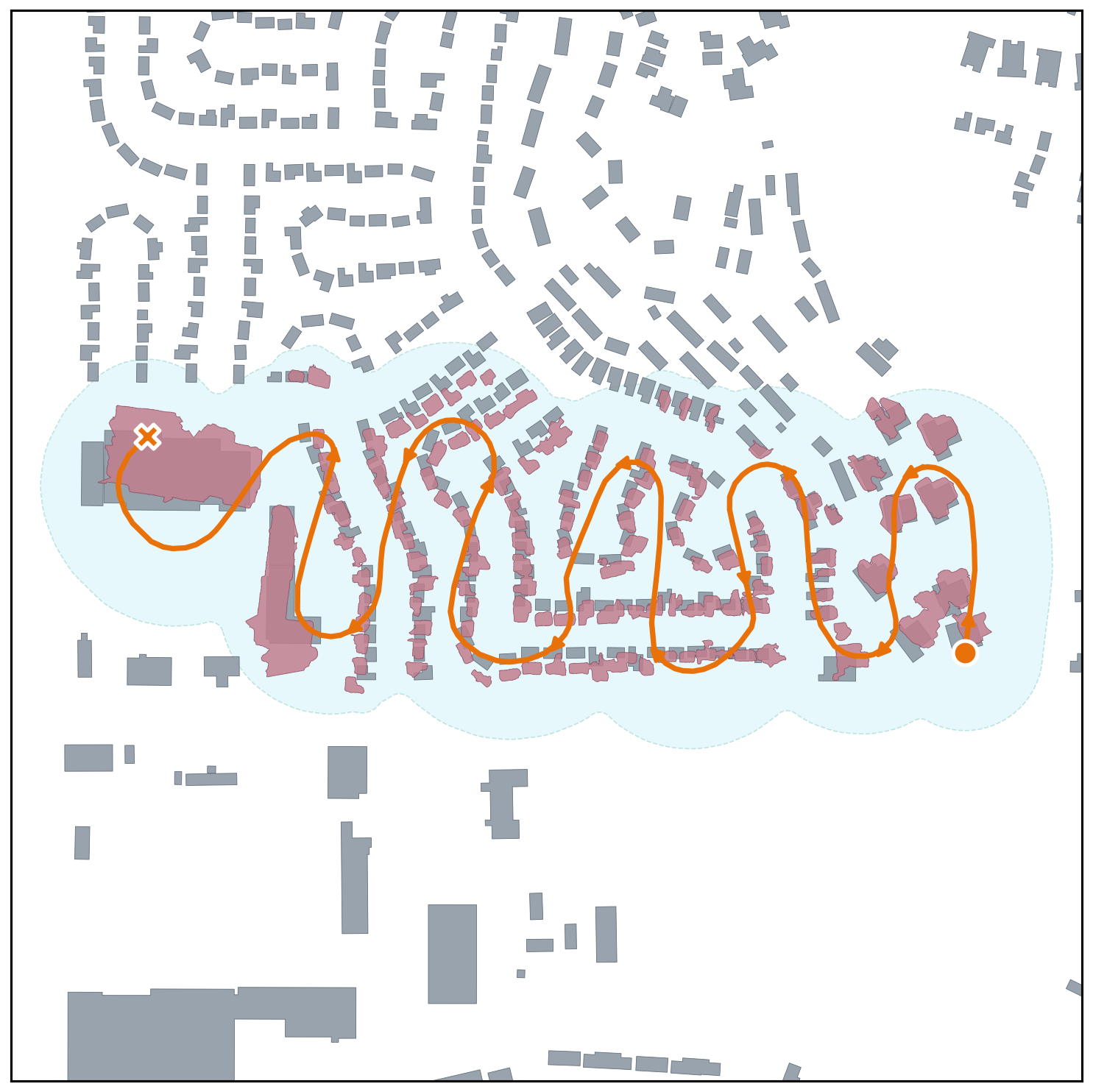}
        \caption{}
        \label{fig:panel_g}
    \end{subfigure}

    \caption{Flight trajectories overlaid on reference building footprint maps. (a)--(g) show flights $1$--$7$ in Table~\ref{tab:flight_summary}.}
    \label{fig:four_panel}
\end{figure*}

\begin{table*}[!t]
\scriptsize
\centering
\caption{Reference-map building count (within each search radius), coarse-stage error, and final localization error across search radii. A trial is counted as successful if the final estimate lies within $100$\,m of ground truth. ``--'' indicates localization failure at flight conclusion.}
\label{tab:flight_buildings_error}
\begin{tabular*}{\linewidth}{@{\extracolsep{\fill}}cccccccccc@{}}
\toprule
\multirow{2}{*}{\textbf{Flight \#}}
& \multicolumn{3}{c}{\textbf{\# Reference Buildings in Search Radius}}
& \multicolumn{3}{c}{\textbf{Coarse-Stage Error (m)}}
& \multicolumn{3}{c}{\textbf{Final Localization Error (m)}} \\
\cmidrule(lr){2-4}\cmidrule(lr){5-7}\cmidrule(lr){8-10}
& \textbf{$R{=}6$ km} & \textbf{$R{=}9$ km} & \textbf{$R{=}12$ km}
& \textbf{$R{=}6$ km} & \textbf{$R{=}9$ km} & \textbf{$R{=}12$ km}
& \textbf{$R{=}6$ km} & \textbf{$R{=}9$ km} & \textbf{$R{=}12$ km} \\
\midrule
$1$ & $18{,}711$ & $51{,}452$ & $99{,}484$ & $22.9$ & $22.9$ & $22.8$ & $7.3$ & $7.3$ & $7.6$ \\
$2$ & $49{,}962$ & $120{,}434$ & $201{,}628$ & $45.0$ & $47.4$ & $47.4$ & $25.4$ & $32.9$ & $31.3$ \\
$3$ & $17{,}987$ & $52{,}290$ & $101{,}521$ & $99.9$ & $102.5$ & $105.3$ & $40.0$ & $38.2$ & $40.7$ \\
$4$ & $84{,}318$ & $178{,}164$ & $272{,}878$ & $27.6$ & $27.4$ & $27.3$ & $20.5$ & $17.7$ & $17.7$ \\
$5$ & $78{,}849$ & $159{,}946$ & $260{,}750$ & $153.7$ & $156.1$ & -- & $22.1$ & $30.3$ & -- \\
$6$ & $90{,}617$ & $167{,}685$ & $277{,}091$ & $52.8$ & $44.2$ & -- & $4.7$ & $4.3$ & -- \\
$7$ & $64{,}662$ & $103{,}742$ & $173{,}956$ & $38.9$ & $36.2$ & $37.8$ & $30.5$ & $25.2$ & $26.8$ \\
\bottomrule
\end{tabular*}
\end{table*}

All pose hypotheses are accumulated in a fixed-resolution grid of $b=100$\,m UTM bins, each contributing a vote weighted by the star score $w_h{=}s$ of its generating match. Writing $g_h$ for the similarity transform of hypothesis $h$ and $x_q$ for the query-map center, the voted position is $\ell_h=g_h(x_q)$ and votes accumulate as
\begin{equation}\label{eq:cg-vote}
V(\kappa)=\sum_{h:\,\beta(\ell_h)=\kappa}w_h,
\end{equation}
with $\kappa$ a bin index and $\beta(\ell)=(\operatorname{round}(\ell_x/b),\operatorname{round}(\ell_y/b))$ the bin of position $\ell$. Geometrically consistent matches reinforce the same bins while incorrect matches disperse, suppressing outliers without explicit hypothesis-level rejection.

In online operation, the internal map grows as keyframes accumulate, while the reference database is fixed for a given search region. Rather than re-matching all internal-map stars at every keyframe, we cache each star's reference matches and re-score only those whose geometry has changed since the previous keyframe; unchanged stars reuse their cached matches. Per-keyframe matching is thus incremental rather than a full re-match against the large reference set.

\subsection{Candidate Selection}

We enforce rotational consistency within each cluster by requiring a minimum number of votes inside an angular tolerance, removing spatially coincident but geometrically inconsistent hypotheses. Each surviving cluster yields a candidate whose similarity transform is the vote-weighted average of its member hypotheses.                                                                                                                                                                                                                        

 For each surviving candidate $c$, applying its similarity transform to the query centroids produces $Q_c$, the predicted positions of observed buildings in the reference frame. We define $R_c$ as the subset of reference buildings that fall within the camera's observable region under $c$---that is, the reference buildings the drone would see if the candidate were correct. Unidirectional nearest-neighbor matching inflates scores for incorrect candidates when a sparse query overlaps a dense reference region, so we instead require mutual nearest-neighbor consistency: given a distance threshold $\tau$, a query--reference pair $(q, r)$ is accepted only if $r$ is the closest reference building to $q$, $q$ is the closest query building to $r$, and $|q - r| < \tau$. We collect all such pairs into $M_c$ and compute the MNN-Jaccard score                                                                                                                                                                                      

  \begin{equation}\label{eq:mnn-jaccard}
  J_c=\frac{|M_c|}{|Q_c|+|R_c|-|M_c|},
  \end{equation}

  \noindent which ranges from $0$ (no mutual matches) to $1$ (every building in both sets is mutually matched). Jaccard normalization further penalizes candidates whose observable region contains many unmatched reference buildings; thus high $J_c$ requires both pairwise agreement and set-level consistency.

 The top $K{=}20$ candidates by weighted vote score are rescored; we select $c^\star=\arg\max_c J_c$ as the coarse localization estimate (coarse-stage error in Table~\ref{tab:flight_buildings_error}); this is neighborhood-scale and may retain residual translation error from map-stitching drift. We then apply a dedicated multi-frame refinement stage that operates on recent raw frames to remove residual translation error.

\subsection{Multi-Frame Refinement and Localization}

The selected coarse candidate $c^\star$ and its similarity transform $(R,t,s)$ retain residual pose error from stitching drift and discretized voting. Refinement corrects this by matching buildings from the last $N$ raw frames (default $N{=}20$) directly to reference centroids. For each frame, buildings are extracted from the segmentation mask, rotated to world coordinates using the per-frame heading derived from the stitching rotation and coarse candidate, and matched to the reference KD-tree via $1$-NN with a distance threshold $d_{\max}=50$\,m.

All matched pairs across frames are pooled into corresponding point sets ${P_i}$ (frame-derived) and ${Q_i}$ (reference). We estimate a residual rigid correction $(\Delta R, \Delta t)$ minimizing $\sum_i \|Q_i - (\Delta R \, P_i + \Delta t)\|^2$ via SVD of the cross-covariance matrix~\cite{arun_least-squares_1987}. The refined last-frame position is $p_{\text{refined}} = \Delta R \, p_{\text{last}} + \Delta t$, jointly correcting residual rotation and translation. Scale is fixed throughout from altitude-derived ground sampling distance.

\section{Experimental Evaluations}\label{sec:evaluations}

\subsection{Dataset and Experimental Setup}

\begin{figure}[!t]
\centering
\includegraphics[width=.99\columnwidth]{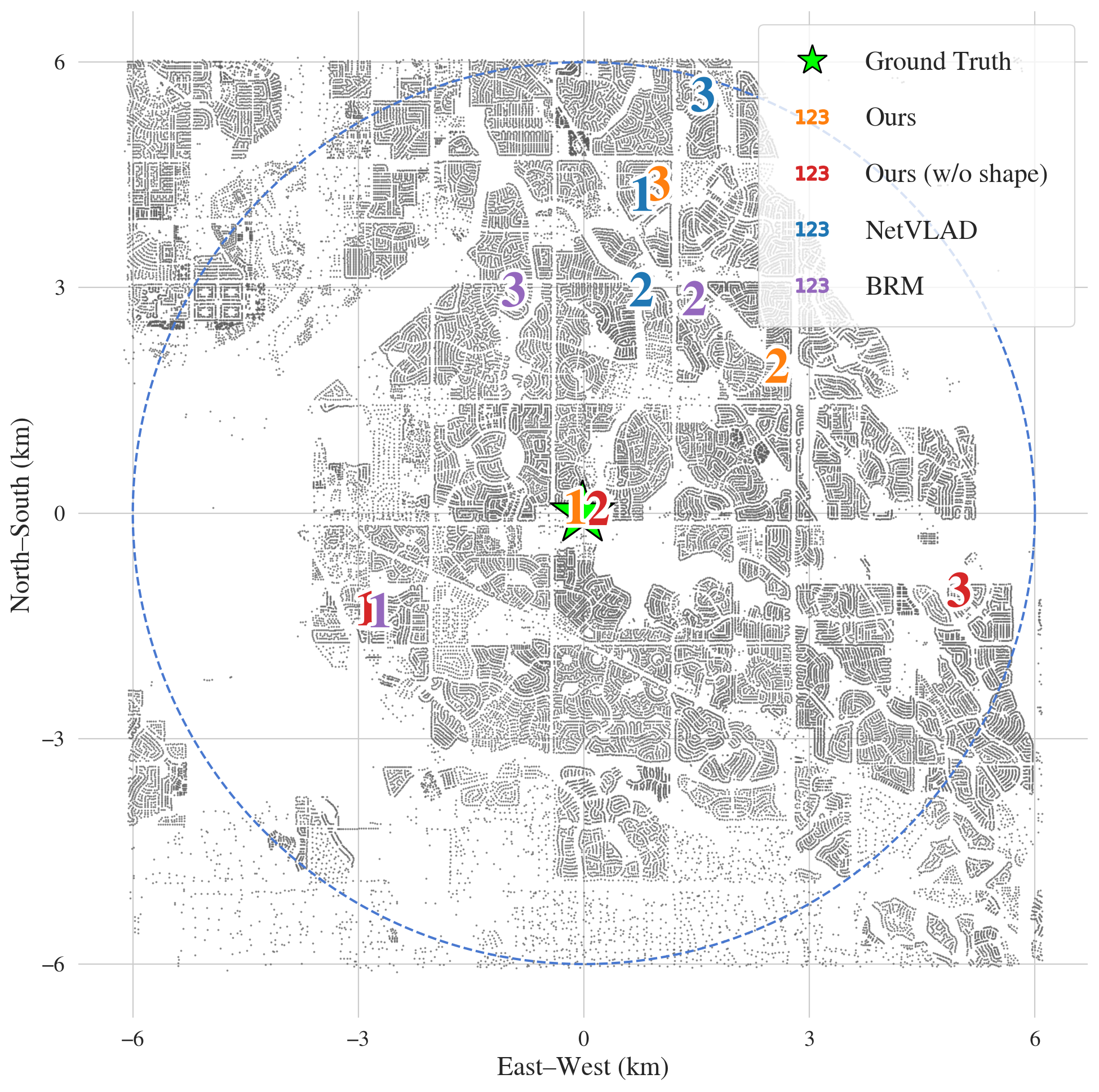}
\caption{Qualitative localization example for Flight~$7$ at the $6$\,km search radius (${\approx}113$\,km$^2$), covering $64,662$ buildings. For each method, the top-$3$ ranked candidates are shown. Our method correctly identifies the true location as its top candidate, while BRM and NetVLAD do not recover the correct location in their top-$3$ candidates. The geometry-only ablation (w/o shape) ranks the correct location lower, illustrating how shape features improve candidate discrimination.}
\label{fig:qualitative_6km_layout}
\end{figure}

We evaluate our algorithm on seven real-world \gls{uav} flights across four municipalities in the Denver metropolitan area (Colorado, USA). Flights span residential, business-district, and suburban areas, and cover $0.39$--$3.18$\,km, $20$--$256$ keyframes, and $97$--$116$\,m altitude above ground level (Table~\ref{tab:flight_summary}, Fig.~\ref{fig:four_panel}). All flights were captured using a DJI Mini 4K platform near midday. Six flights were recorded in fair, snow-free conditions, while Flight~$7$ was flown several days after snowfall with moderate residual snow cover, introducing ground-surface artifacts absent from the other flights. Ground-truth trajectories are derived from onboard consumer-grade GPS logs, which contribute some horizontal uncertainty to all reported errors.

Reference maps are built from Microsoft Building Footprints with search radii \(R\in\{6,9,12\}\,\mathrm{km}\), corresponding to approximately $113$, $254$, and $452$\,km$^2$, respectively. These maps contain approximately $18{,}000$--$277{,}000$ reference buildings (Table~\ref{tab:flight_buildings_error}). On consumer hardware (NVIDIA RTX $5070$), measured per-query runtime for segmentation inference, incremental candidate matching and scoring with online caching, and frame refinement is $0.4$--$1.2$\,s at $6$\,km, $0.6$--$1.3$\,s at $9$\,km, and $0.7$--$1.5$\,s at $12$\,km; times exclude one-time offline preprocessing of reference descriptors.

\textbf{Evaluation metrics.} All methods are evaluated over search radii of $1$, $3$, $6$, $9$, and $12$\,km. A prediction is considered correct if the estimated position is within $100$\,m of ground truth. This $100$\,m acceptance radius is a deliberate design choice matching the $b{=}100$\,m voting-grid bin used in candidate generation (Sec.~\ref{sec:candidate-generation}); it bounds coarse-retrieval success at metropolitan scale, whereas achieved accuracy is finer (final localization errors $4.3$--$40.7$\,m, Table~\ref{tab:flight_buildings_error}). We report Top-$1$ localization error and Recall@$K$ (R@$K$), where R@$K$ is the fraction of flights with at least one correct hypothesis in the top-$K$ ranked candidates.

For partial-information analysis, each flight is re-evaluated over temporal prefixes (the first $k$ keyframes, $k{=}1,\ldots,N$). We report first-success and stable-success keyframe counts: the earliest prefix at which Top-$1$ error is ${\le}100$\,m, and the earliest prefix at which it is ${\le}100$\,m and remains so for all subsequent prefixes of the same flight.

\subsection{Compared Methods}

\begin{table}[!t]
\scriptsize
\centering
\caption{Comparison with baselines and ablation across search radii with a $100$\,m correctness threshold. Entries are Recall@$K$ values reported as successful flights out of $7$. Best results per column are shown in bold.}
\label{tab:baseline_comparison}
\begin{tabular*}{\linewidth}{@{\extracolsep{\fill}}llccccc@{}}
\toprule
\textbf{Method} & \textbf{Metric} & \makecell{\textbf{$R{=}$} \\ \textbf{$1$km}} & \makecell{\textbf{$R{=}$} \\ \textbf{$3$km}} & \makecell{\textbf{$R{=}$} \\ \textbf{$6$km}} & \makecell{\textbf{$R{=}$} \\ \textbf{$9$km}} & \makecell{\textbf{$R{=}$} \\ \textbf{$12$km}} \\
\midrule
NetVLAD & R@$1$  & $5/7$ & $2/7$ & $0/7$ & $0/7$ & $0/7$ \\
NetVLAD & R@$5$  & $\mathbf{7/7}$ & $3/7$ & $0/7$ & $0/7$ & $0/7$ \\
NetVLAD & R@$10$ & $\mathbf{7/7}$ & $3/7$ & $0/7$ & $0/7$ & $0/7$ \\
BRM     & R@$1$  & $6/7$ & $1/7$ & $1/7$ & $0/7$ & $0/7$ \\
BRM     & R@$5$  & $6/7$ & $2/7$ & $1/7$ & $0/7$ & $0/7$ \\
BRM     & R@$10$ & $6/7$ & $3/7$ & $1/7$ & $0/7$ & $0/7$ \\
Ours (w/o shape) & R@$1$  & $5/7$ & $5/7$ & $4/7$ & $3/7$ & $3/7$ \\
Ours (w/o shape) & R@$5$  & $6/7$ & $6/7$ & $5/7$ & $3/7$ & $3/7$ \\
Ours (w/o shape) & R@$10$ & $6/7$ & $6/7$ & $6/7$ & $3/7$ & $3/7$ \\
\midrule
\textbf{Ours} & R@$1$  & $\mathbf{7/7}$ & $\mathbf{7/7}$ & $\mathbf{7/7}$ & $\mathbf{7/7}$ & $\mathbf{5/7}$ \\
\textbf{Ours} & R@$5$  & $\mathbf{7/7}$ & $\mathbf{7/7}$ & $\mathbf{7/7}$ & $\mathbf{7/7}$ & $\mathbf{5/7}$ \\
\textbf{Ours} & R@$10$ & $\mathbf{7/7}$ & $\mathbf{7/7}$ & $\mathbf{7/7}$ & $\mathbf{7/7}$ & $\mathbf{5/7}$ \\
\bottomrule
\end{tabular*}
\end{table}

We compare against two representative baselines: BRM~\cite{choi_brm_2020} and NetVLAD~\cite{arandjelovic_netvlad_2018}. For BRM, we implement the building-ratio descriptor using building-pixel coverage in concentric circular regions derived from the camera footprint, yielding a rotation-invariant 3-element ratio vector at each hypothesized location. Reference ratio maps are precomputed from rasterized building footprints. BRM inference follows the original threshold-based candidate generation and motion-constrained propagation in~\cite{choi_brm_2020}: candidates are selected by applying an $L_1$ residual threshold between the query descriptor and the reference ratio maps, then propagated across frames using odometry constraints. To report results under a retrieval-based protocol, we rank only the surviving BRM candidates at the end of the flight by their accumulated $L_1$ residual and evaluate success based on the top-$K$ ranked hypotheses.

For NetVLAD, we use a dual-branch ResNet-$18$ backbone to handle the cross-view domain gap between UAV imagery and reference tiles, with $4096$-D descriptors trained on UAV-VisLoc~\cite{xu2024uavvisloclargescaledatasetuav} using triplet loss. NetVLAD is evaluated as a retrieval baseline by scoring a discretized set of reference tiles within the search radius and accumulating multi-frame $L_2$ distances over odometry-propagated hypotheses.

In coarse localization, our method and BRM are evaluated without known compass heading. The adapted NetVLAD baseline uses compass-heading-assisted multi-frame propagation because its learned descriptors are not rotation-invariant; heading is required to align query frames to reference tiles before descriptor extraction.

To assess how much per-building shape features contribute beyond the base geometric formulation of Li et al.~\cite{li_localization_2020}, we evaluate a geometry-only ablation of our pipeline. This variant, labeled ``Ours (w/o shape)'' in Table~\ref{tab:baseline_comparison}, replaces the full $32$-dimensional star descriptor with an $8$-dimensional version following Li et al.'s star descriptor exactly: each triangle is represented by its area $A$ and squared perimeter $l{=}L^2$, concatenated for the central and three neighboring triangles. All other stages remain identical, so any difference reflects the descriptor change alone.

\subsection{Benchmark Results}\label{sec:benchmark_results}

Table~\ref{tab:baseline_comparison} shows the primary benchmark result: our method localizes $7/7$ flights at $6$\,km and $9$\,km search radii, while BRM and NetVLAD drop to $0/7$ at $9$\,km. At $12$\,km, our method localizes $71.4\%$ ($5/7$) of flights, while both baselines remain at $0/7$. Across all three radii, this corresponds to $90.5\%$ success overall ($19/21$ trials).
To our knowledge, this is the largest search area reported to date for localizing with real-world drone imagery under a cross-view \gls{uav}-to-map protocol. A qualitative example at $6$\,km is shown in Fig.~\ref{fig:qualitative_6km_layout}.

The geometry-only ablation (Table~\ref{tab:baseline_comparison}, ``Ours (w/o shape)'') consistently underperforms the full method, with the gap widening at larger search radii. R@$1$ degrades from $7/7$ to $5/7$ at $1$ and $3$\,km, to $4/7$ at $6$\,km, and to $3/7$ at $9$ and $12$\,km (compared to $7/7$, $7/7$, and $5/7$ for the full method). This increasing gap reflects the growing disambiguation challenge: as the candidate region expands, urban areas contain many geometrically similar building configurations that cannot be resolved by triangle geometry alone, and per-building shape features provide the additional discriminability needed to rank the correct candidate first. Notably, the ablation still outperforms both BRM and NetVLAD at all large search radii, confirming that Delaunay-based geometric matching is the primary source of success and that shape augmentation is an important but additive refinement.

Table~\ref{tab:flight_buildings_error} reports our end-to-end error statistics along with reference-map scale. At $6$\,km and $9$\,km, all seven flights localize successfully, with final error ranges of $4.7$--$40.0$\,m and $4.3$--$38.2$\,m, respectively; at $12$\,km, successful runs span $7.6$--$40.7$\,m. Larger residuals typically occur in regions with sparse or spatially uniform building distributions, where fewer distinctive geometric cues are available for refinement alignment. Under our evaluation criterion (Top-$1$ error \(\le 100\)\,m), these are still correct large-area recoveries, and they occur at scales where both baselines fail entirely.

\begin{figure}[!t]
\centering
\includegraphics[width=1.\columnwidth]{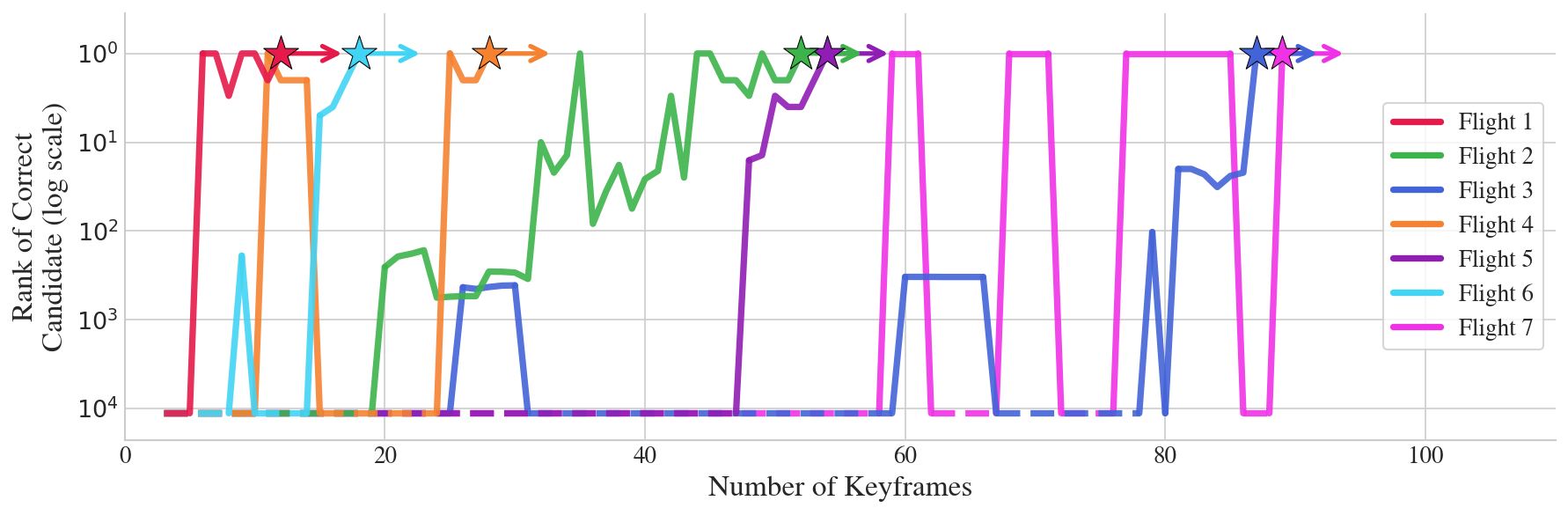}
\caption{Progressive coarse localization at $R{=}6$\,km. Each trace shows the best rank among the ground-truth bin and its 8-connected neighboring bins among $11{,}310$ candidates as keyframes accumulate from $1$ to $N$. Only rank-$1$ constitutes coarse-stage success. Star markers denote each flight's first stable rank-$1$ point. Fluctuations arise from false-positive or missed building detections that perturb the internal Delaunay topology and alter scoring.}
\vspace*{-0.5em}
\label{fig:convergence}
\end{figure}

\subsection{In-Flight Localization Under Partial Information}\label{sec:partial_info_results}

We next examine a key operational concern: whether reliable coarse localization can be achieved mid-flight, before the full trajectory is available. For each flight, we re-run the coarse localization pipeline at search radii \(R\in\{6,9,12\}\,\mathrm{km}\) on temporal prefixes of $1$ to $N$ keyframes (excluding the final multi-frame refinement stage). At each prefix length, we run the coarse pipeline (through MNN-Jaccard rescoring) and rank all candidate grid bins by their score, assigning unvoted bins a tied last-place rank. The reported rank at each prefix length is the best rank among the ground-truth bin and its 8-connected neighboring bins, accounting for the fact that stitching drift can place a near-correct candidate in an adjacent bin.

Fig.~\ref{fig:convergence} visualizes this progression at $R{=}6$\,km. Localization rank can fluctuate early under limited evidence as buildings enter the observed set, altering the Delaunay topology and scoring, then stabilize as additional observations accumulate. Flight~$7$ illustrates the effect of adverse conditions on convergence speed: after briefly achieving rank-$1$, the correct candidate loses its ranking as later keyframes introduce snow-covered ground patches that the segmentation network misclassifies as building footprints. These spurious detections inject phantom vertices into the observed Delaunay triangulation, corrupting the star descriptors of neighboring buildings and delaying stable localization, though the correct candidate is ultimately recovered. At $6$\,km, mean first-success and stable-success times are $38.6$ and $48.6$ keyframes, respectively; at $9$\,km, $42.7$ and $51.6$; and at $12$\,km, $64.2$ and $75.4$ over the five successful flights. The first stable rank-$1$ points provide a trajectory-level view of when reliable coarse in-flight localization is reached before refinement.

\section{Limitations and Future Work}

Our design prioritizes large-area candidate-space reduction before fine localization, and several limitations remain. First, residual localization error remains nontrivial: successful trials fall within $100$\,m, but final errors span $4.3$--$40.7$\,m. Multi-frame refinement removes much of the coarse error but is bounded by stitched-map quality: pose-graph drift perturbs centroids and segmentation errors reduce correspondences for the SVD correction. Second, scalability degrades with search area and urban density as candidate count, descriptor comparisons, and voting load grow, and the star-descriptor and pruning design's robustness--efficiency sensitivity is not yet systematically characterized. Third, the building-only reference limits available constraints and is sensitive to footprint incompleteness: reliability drops when footprints are missing or displaced enough to alter local Delaunay topology. Fourth, the evaluation assumes nadir imagery under mostly daytime, clear-visibility conditions with accurate inter-frame alignment, so end-to-end performance is not fully appearance-invariant and depends on reliable query-side building extraction under challenging conditions.

Future work will (i) target meter-level accuracy through stronger final alignment, lower-drift state estimation, and sensor fusion; (ii) extend the map beyond buildings to other stable GIS layers (e.g., road networks, water boundaries); (iii) run systematic ablations of star-descriptor components and candidate-pruning strategies to quantify accuracy--efficiency tradeoffs; and (iv) evaluate and adapt the pipeline under illumination, seasonal, and adverse-weather variation.

\section{Conclusion}%

We presented a vision-based \gls{uav} localization system that matches observed building geometry to vector building-footprint maps. 
Across seven flights, our system achieves $100\%$ localization success at $6$\,km and $9$\,km search radii and $71.4\%$ ($5/7$) at $12$\,km, corresponding to $90.5\%$ ($19/21$) successful trials overall. 
These results show that geometric structure supports reliable large-area localization under weak priors (no compass heading, no raster reference imagery). Compared with BRM~\cite{choi_brm_2020} and NetVLAD~\cite{arandjelovic_netvlad_2018} across $1$--$12$\,km search radii (up to ${\approx}452$\,km$^2$), our method maintains substantially stronger recall as search area expands. 

\section*{Disclaimer}

The views expressed in this paper are those of the authors and do not reflect the official guidance or position of the United States Government, the Department of Defense, the United States Air Force, or the United States Space Force.

\balance %

\bibliographystyle{IEEEtran}
\bibliography{refs}

\end{document}